\documentclass[10pt,twocolumn,letterpaper]{article}

\usepackage[pagenumbers]{cvpr} 

\definecolor{cvprblue}{rgb}{0.21,0.49,0.74}
\usepackage[pagebackref,breaklinks,colorlinks,allcolors=cvprblue]{hyperref}

\def\paperID{3203} 
\def\confName{CVPR}
\def\confYear{2025}

\title{Introducing Visual Perception Token into Multimodal Large Language Model}

\author{Runpeng Yu\textsuperscript{*}, Xinyin Ma\textsuperscript{*} and Xinchao Wang\textsuperscript{\textdagger}\\
National University of Singapore\\
{\tt\small \{r.yu,maxinyin\}@u.nus.edu and xinchao@nus.edu.sg}
}
\usepackage{bm}

\usepackage{subcaption}

\usepackage{multirow} 
\usepackage{makecell}

\usepackage{tikz}

\definecolor{myred}{rgb}{1.0, 0.62, 0.61}
\definecolor{mygreen}{RGB}{8, 147, 146}
\definecolor{myorange}{RGB}{233, 159, 105}
\definecolor{mypink}{RGB}{207, 89, 126}

\usepackage{array} 

\usepackage{setspace}
\usepackage{tocloft}

\usepackage{amssymb} 
\usepackage{pifont}
\definecolor{fullgreen}{rgb}{0.502, 0.788, 0.643}
\definecolor{fullred}{rgb}{0.800, 0.447, 0.541}
\newcommand{\cmark}{\textcolor{fullgreen}{\textbf{\checkmark}}}
\newcommand{\xmark}{\textcolor{fullred}{\textbf{\ding{55}}}}
\definecolor{lightgreen}{RGB}{225, 239, 217}   
\definecolor{lightblue}{RGB}{203, 220, 235}   
\definecolor{fullgray}{RGB}{219, 223, 234}   
\definecolor{fullpurple}{RGB}{205, 193, 255}
\definecolor{darkred}{RGB}{204, 114, 138}
\definecolor{darkpurple}{RGB}{171, 151, 255}
\definecolor{darkgray}{RGB}{114, 114, 114}

\newcommand{\rsta}{\colorbox{fullgray}{\strut st}$\,$}
\newcommand{\rstb}{\colorbox{lightblue}{\strut $x_{\text{min}}$}$\,$}
\newcommand{\rstc}{\colorbox{lightblue}{\strut $y_{\text{min}}$}$\,$}
\newcommand{\rstd}{\colorbox{lightblue}{\strut $x_{\text{max}}$}$\,$}
\newcommand{\rste}{\colorbox{lightblue}{\strut $y_{\text{max}}$}$\,$}
\newcommand{\rstf}{\colorbox{fullgray}{\strut  ed}}

\newcommand{\dftb}{\colorbox{lightblue}{\strut  DINO\_Ctrl}$\,$}

\usepackage{colortbl}

\usepackage{soul}
\setul{0.1ex}{0.3ex}
\usepackage{enumitem}

\usepackage{listings}
\usepackage{tcolorbox}
\begin{document}
\twocolumn[{%
\renewcommand\twocolumn[1][]{#1}%
\maketitle
\begin{center}
    \centering
    \captionsetup{type=figure}
    \begin{subfigure}[b]{0.73\linewidth}
        \includegraphics[page=1, trim=0in 0in 1.9in 0in, clip, width=\textwidth]{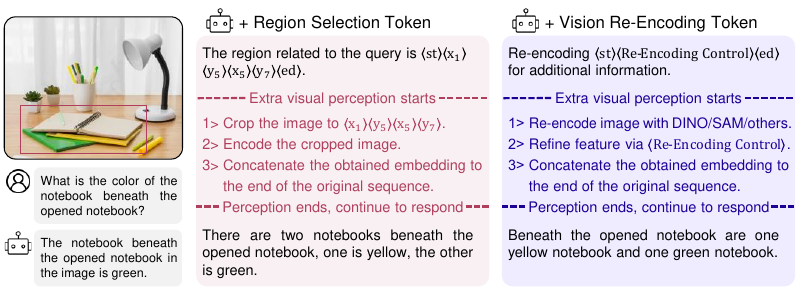}
        \caption{Response process of an MLLM equipped with Visual Perception Tokens.}
        \label{fig:head:task}
    \end{subfigure}
    \hfill
    \begin{subfigure}[b]{0.26\linewidth}
        \includegraphics[page=1, trim=5.4in 0in 0in 0in, clip, width=\textwidth]{fig/head2.pdf}
        \caption{Average Performance.}
        \label{fig:head:performance}
    \end{subfigure}
    \vspace{-0.5em}
    \caption{The Visual Perception Token aids MLLMs by triggering and controlling additional visual perception processes. \cref{fig:head:task} illustrates the response process of MLLMs equipped with the Region Selection Token or Vision Re-Encoding Token. The regions marked in the image are selected by the Region Selection Token. \cref{fig:head:performance} presents the average performance of a 2B model with the Visual Perception Token across various VQA tasks (higher is better, with 1 being the maximum).}
    \label{fig:head}
\end{center}%
}]
\let\thefootnote\relax\footnotetext{\textsuperscript{*} Equal Contribution; \textsuperscript{\textdagger} Corresponding Author.}

\begin{abstract}
To utilize visual information, Multimodal Large Language Model (MLLM) relies  on the perception process of its vision encoder. The completeness and accuracy of visual perception significantly influence the precision of spatial reasoning, fine-grained understanding, and other tasks. 
However, MLLM still lacks the autonomous capability to control its own visual perception processes, for example, selectively reviewing specific regions of an image or focusing on information related to specific object categories.
In this work, we propose the concept of Visual Perception Token, aiming to empower MLLM with a mechanism to control its visual perception processes. We design two types of Visual Perception Tokens, termed the Region Selection Token and the Vision Re-Encoding Token. MLLMs autonomously generate these tokens, just as they generate text, and use them to trigger additional visual perception actions. The Region Selection Token explicitly identifies specific regions in an image that require further perception, while the Vision Re-Encoding Token uses its hidden states as control signals to guide additional visual perception processes.
Extensive experiments demonstrate the advantages of these tokens in handling spatial reasoning, improving fine-grained understanding, and other tasks. On average, the introduction of Visual Perception Tokens improves the performance of a 2B model by 30.9\%, increasing its score from 0.572 to 0.749, and even outperforms a 7B parameter model by 20.0\% (from 0.624). Please check out our repo \href{https://github.com/yu-rp/VisualPerceptionToken}{here}.
\end{abstract}    
\section{Introduction}
\label{sec:intro}

Multimodal Large Language Model (MLLM) depend on the perception capabilities of their vision encoder to process and utilize visual information. During this process, MLLM utilizes a vision encoder and a projector to embed visual information into the language space. The quality of Visual Perception determines whether MLLMs can accurately distinguish objects in an image~\cite{mot}, whether MLLMs can rely on visual information to answer questions instead of generating textual hallucinations~\cite{leng2024mitigating}, and whether MLLMs can perform precise reasoning about spatial relationships~\cite{cheng2024spatialrgpt}, among other tasks.
While current MLLM systems demonstrate strong capabilities in visual information understanding~\cite{gpt4mllm,gemini,Qwen2VL,llava15}, they lack the ability to autonomously control their Visual Perception processes. Instead, these systems depend on manually designed pipelines to perform specific image annotations or visual features enhancement~\cite{yu2024api,mot}. 

In this work, we explore the task of enabling MLLMs to autonomously control their Visual Perception processes. 
Previously, MLLM-based agents and MLLMs equipped with function-calling or tool-use capabilities can be considered as having the ability to control subsequent tasks. They utilize the output of the LLM as arguments for subsequent functions or tool use. However, such control information is confined to the natural language space.  
The advantage of control signals in the natural language space lies in their interpretability, clear supervision signals, and ease of training data construction. However, these signals are constrained by specific formats. Additionally, natural language inherently contains redundancy, leading to efficiency issues.  
In this work, we aim to explore control signals beyond the natural language space. However, we also require that these signals remain naturally compatible with the next-token prediction paradigm of LLMs. 
To address this, we propose the concept of ``Visual Perception Tokens''. These tokens are integrated into the MLLM vocabulary and can be generated by the MLLM through next-token prediction, similar to natural language generation. These tokens do not correspond to specific words or characters in natural language; instead, their primary function is to trigger additional Visual Perception processes and convey control information for these processes. 

\setulcolor{fullred!50} 
We designed two types of Visual Perception Tokens. The first type is \ul{Region Selection Token}, which instruct the MLLM to crop the input image and encode again the important regions relevant to the query using the vision encoder. 
\setulcolor{fullpurple!70} 
The second type is the \ul{Vision Re-Encoding Token}, which signals the model to input the image into (additional) vision encoder and use the resulting vision features to supplement the original MLLM's vision features. 
A projector takes both the additional vision features and the hidden state of the \ul{Vision Re-Encoding Token} as inputs, enabling fine-grained control beyond merely triggering the vision encoder.
In this work, we explore using an additional DINO model, a SAM model, or the model’s original vision branch as the vision encoder.
During the generation process, if the MLLM outputs any Visual Perception Token, the corresponding additional perception process is triggered, and the extra embedding sequence derived from the image is concatenated to the original LLM input. The LLM then continues generating the response in the form of next-token prediction. \cref{fig:head:task} illustrates the VQA process incorporating visual perception tokens. \cref{fig:forward} provides a more detailed depiction of how visual perception tokens are generated by the MLLM and how they are utilized to control the visual perception process.

To train the MLLM to use Visual Perception Tokens, we constructed the Visual Perception Token training dataset, which includes 829k samples spanning four task categories: General VQA, Fine-Grained VQA, Spatial Reasoning, and Text/OCR-Related VQA. Experiments demonstrated that Visual Perception Token significantly enhances the MLLM's ability to autonomously control and refine its visual perception.

Our contributions can be summarized as follows:
\begin{enumerate}[labelwidth=0pt,labelsep=3pt,itemindent=0em]
\item We explored a novel task: enabling MLLMs to autonomously control their visual perception process. 
\item We designed two types of Visual Perception Tokens: one enabling the MLLM to select regions of interest, and another allowing the MLLM to incorporate additional vision features and control the final embeddings input to the language model.
\item Experimental results demonstrate the effectiveness of our approach. On tasks such as Spatial Reasoning and Fine-Grained VQA, models equipped with Visual Perception Tokens achieved performance improvements of 34.6\% and 32.7\% over the 7B baseline model, respectively.
\end{enumerate}
\section{Related Work}

\subsection{Visual Prompting}
From a technical perspective, our approach can also be regarded as a learnable visual prompting method.
Visual prompting is a key technique in vision models, especially for segmentation tasks \cite{sam,vp_3,vp_2}. It uses a prompt encoder to interpret manual annotations, such as points and masks, to control segmentation granularity and assist instance selection. Recent advancements show that LVLMs can interpret visual cues like circles and color masks in a zero-shot manner without an additional encoder \cite{vp_clip_1,vp_clip_2}. 
Building on this, \cite{som} and \cite{fgvp} have utilized segmentation model-generated masks as visual prompts, enhancing LVLM performance in segmentation and grounding. However, these methods are query-agnostic, while VQA tasks require adapting visual information based on the query. \cite{yu2024api} addresses this by using an auxiliary VLM to generate query-specific visual prompts, overlaying attention maps to guide focus on relevant regions.

Built on these learning-free methods, \cite{shao2024visual} trains the MLLM to output bounding boxes of important regions. The image is then cropped and re-input for inference, creating a "crop and re-input" CoT process. Compared to \cite{shao2024visual}, our design of Region Selection Tokens does not rely on bounding box information in the natural language space but instead uses specialized tokens to indicate the location of important regions. This design simplifies training and mitigates the issue of MLLMs having difficulty aligning the image coordinate system with coordinates described in natural language.

\subsection{Visual Perception in MLLM}
Currently, MLLMs have developed the ability to extract visual information, enabling them to be used not only for general image-based question answering and chat~\cite{llava,llava15,cogvlm,gpt4v,gemini,Qwen2VL,Qwen-VL,ma2023llmpruner}, but also for applications in 3D understanding~\cite{hong20233dllminjecting3dworld,zhu2024llava,cvpr23_3dclr}, video analysis~\cite{damonlpsg2024videollama2,damonlpsg2023videollama,tang2024salmonn,sun2024videosalmonn}, domain-specific image question answering~\cite{li2023llavamed}.

Our objective is to further enhance the visual perception capabilities of MLLMs using Visual Perception Tokens. In addition to the aforementioned visual prompting techniques, there are various other research directions focused on improving MLLM visual perception performance.

\cite{mot} identifies the ambiguity within CLIP's features as a limitation that adversely affects MLLMs' visual perception performance. To overcome this, it proposes incorporating additional DINOv2~\cite{oquab2024dinov} encoders to boost the visual perception capabilities of MLLMs. 
\cite{wang2024pictureworththousandwords} observed that even with access to image data, MLLMs can sometimes base their responses on textual information and hallucinations instead of directly leveraging the visual content. To address this, \cite{wang2024pictureworththousandwords} suggests adding image captions to improve visual perception accuracy.
In the specific task of spatial reasoning, visual perception focuses on inferring spatial relationships between objects. To enhance a model's spatial reasoning capabilities, \cite{cheng2024spatialrgpt} proposed using depth maps of scenes as additional input. Moreover, \cite{cvpr23_3dclr} has explored how to leverage multiview scene information to improve the visual perception of MLLMs. To transform multiview information into a suitable input format, \cite{cvpr23_3dclr} employed a neural field representation. Similarly, to address the limitations of 2D images in physical world reasoning, \cite{Chen_2024_CVPR,zhu2024llava} have investigated MLLM visual perception methods based on 3D data. In this case, the input to MLLMs can include additional 3D point cloud features~\cite{hong20233dllminjecting3dworld} and 3D scene graph features~\cite{conceptgraphs}.

The above approaches aim to provide MLLMs with better visual inputs to enhance their visual perception abilities. In contrast, our approach features iterative perception, allowing MLLMs to provide feedback on the perception process, conduct multiple rounds of visual perception, and exercise control over the visual perception process.
\section{Visual Perception Token}

We add two types of extra Visual Perception Tokens to the original MLLM vocabulary. \cref{tab:compare} provides a comparison of their key attributes. Previously, MLLM  could only generate rationales and answers in natural language. However, with the addition of Visual Perception Tokens, the MLLM can now output valuable information in non-natural language form. These Visual Perception Tokens serve primarily as triggers; when a Visual Perception Token is output, the MLLM initiates additional visual perception processes.

\begin{figure}[t]
  \centering
  \includegraphics[width=0.7\linewidth]{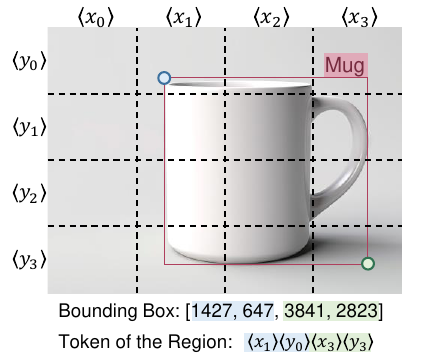}
   \caption{The relationship between the Region Selection Tokens we used and a precise bbox. Region Selection Token uses the cells containing the top-left and bottom-right corners to describe the approximate location of the region. In this example, the image is evenly divided into $4\times 4$ cells. In our main experiment, we divide images into $8\times 8$ cells.}
   \label{fig:bbox_token}
\end{figure}

\begin{table}[t]
    \centering
    \resizebox{\linewidth}{!}{%
    \begin{tabular}{@{}lcc@{}}
        \toprule
         & \textbf{Region Selection Token}& \textbf{Vision Re-Encoding Token}  \\ 
        \midrule
        Format  &  \rsta\rstb\rstc\rstd\rste\rstf
        & \rsta\dftb\rstf \\
        Function  & Crop the Image & Input Vision Feature \rule{0pt}{13pt}\\
        Encoder  & Original Vision Encoder & DINO/SAM/...  \\
        Informative Hidden States   &\cellcolor{fullred!10} \xmark &\cellcolor{fullgreen!10} \cmark \\
        Supervision Signal  &\cellcolor{fullgreen!10} \cmark &\cellcolor{fullred!10} \xmark  \\
        Clear Semantic Meaning  &\cellcolor{fullgreen!10} \cmark &\cellcolor{fullred!10} \xmark  \\ \bottomrule
    \end{tabular}%
    }
    \caption{Comparison between the Region Selection Token and Vision Re-Encoding Token.}
    \label{tab:compare}
\end{table}

\subsection{Region Selection Tokens} 
\label{sec:31}

Please see \cref{fig:bbox_token} for the relationship between the Region Selection Tokens we used and a precise bbox and see \cref{fig:head:task} for an example when MLLM response with Region Selection Tokens. 

Each group of Region Selection Tokens represents a bounding box (bbox). After the MLLM outputs a group of Region Selection Tokens, the original image will be cropped according to the bbox, preserving only the regions relevant to the query, and then be re-input into the MLLM. This ``crop and re-input'' approach enhances visual perception performance by directly increasing resolution. Although simple, it has been shown to be highly effective~\cite{llava15,shao2024visual}. For example, in document understanding and OCR-related tasks, the region containing the target text is often small, while the original document or image may be large and might even need to be downsized to fit the model input. In such cases, recognizing the query-relevant text becomes very challenging. However, cropping the image and then upsizing the relevant region before inputting it into the model significantly improves task performance.

Given an $h \times w$ image and a rectangular region $R$, a bounding box $[x_{\text{min}}, y_{\text{min}}, x_{\text{max}}, y_{\text{max}}]$ can precisely describe the position of region $R$, where $(x_{\text{min}}, y_{\text{min}})$ and $(x_{\text{max}}, y_{\text{max}})$ represent the coordinates of the top-left and bottom-right pixels of region $R$, respectively. However, precise pixel coordinates are not necessary for VQA tasks. Previous research also shows that MLLMs struggle to interpret and generate precise bounding boxes~\cite{gpt4early}, especially when the input image resolution is not fixed, as is often the case in practice. In such cases, MLLMs struggle to infer the original image resolution from the patchified image embeddings, making it difficult to generate valid bounding boxes.
Therefore, instead of using exact pixel coordinates to describe a region's bounding box, we describe only the approximate location of the region. We divide the $h \times w$ image evenly into a grid of $k \times k$ rectangular cells, with each cell sized $\frac{h}{k} \times \frac{w}{k}$. Each cell can be indexed by its row and column, with the top-left cell indexed as $(0, 0)$ and the top-right cell as $(k-1, 0)$. We use the indices of the cells containing the top-left and bottom-right pixels of region $R$ to describe its location. In our implementation, we set $k=8$.

A group of Region Selection Tokens includes six consecutive tokens: a $<$Region\_Selection\_Start$>$ token, two tokens representing the index of the top-left cell, two tokens representing the index of the bottom-right cell, and a $<$Region\_Selection\_End$>$ token.
To enable the model to better distinguish between horizontal and vertical coordinates, we added $2k$ tokens, $<\!\!x_0\!\!>,\cdots,<\!\!x_{k-1}\!\!>$ and $<\!\!y_0\!\!>,\cdots,<\!\!y_{k-1}\!\!>$, specifically for indicating cell indices. For example, the top-left cell is represented as $<\!\!x_0\!\!><\!\!y_0\!\!>$ and the top-right cell as $<\!\!x_{k-1}\!\!><\!\!y_0\!\!>$.
Given a group of Region Selection Tokens, the pixel coordinates for the top-left and bottom-right corners of the region to be cropped are calculated as $([x_{\text{min}} \times \frac{w}{k}, y_{\text{min}} \times \frac{h}{k})$ and $((x_{\text{max}}+1) \times \frac{w}{k}, (y_{\text{max}}+1) \times \frac{h}{k}])$, respectively. 

Based on this design, Region Selection Tokens have clear and interpretable semantics, indicating specific locations within the image. During training, Region Selection Tokens are learned using the next token prediction loss.

\subsection{Vision Re-Encoding Tokens} 
Please see \cref{fig:head:task} for an example when MLLM response with Vision Re-Encoding Tokens. 

The Vision Re-Encoding Tokens trigger an additional vision encoder, such as DINO, to re-encode the original image, with the resulting vision features processed by a projector before being input into the MLLM. 

Each set of Vision Re-Encoding Tokens consists of three tokens: $<$Re-Encode\_Start$>$ token, $<$Re-Encoding\_Control$>$, and a $<$Re-Encode\_End$>$ token. The hidden state of the $<$Re-Encoding\_Control$>$ token token together with the vision features is input into the projector, allowing further control over the final embedding sequence fed into the LLM. 
To enable $<$Re-Encoding\_Control$>$ to freely convey any control information, we do not calculate the loss at the $<$Re-Encoding\_Control$>$ token during training. As a result, the MLLM's output at the $<$Re-Encoding\_Control$>$ token can be arbitrary. We only require its hidden states to be informative. From this perspective, $<$Re-Encoding\_Control$>$ does not convey clear, interpretable semantics and cannot be decoded into a specific word or phrase. This is also the reason that additional $<$Re-Encode\_Start$>$ and $<$Re-Encode\_End$>$ tokens are used to mark the presence of $<$Re-Encoding\_Control$>$.

\textbf{Mask Modeling}. To enhance the performance of $<$Re-Encoding\_Control$>$ token, we adopt a training approach similar to Masked Language Modeling. During training, 50\% of the samples containing the Vision Re-Encoding Token undergo extra masking.  
We modify the attention mask for these samples, ensuring that the tokens corresponding to the answer in the dialogue can only access the $<$Re-Encoding\_Control$>$ token while being restricted from accessing the tokens corresponding to the original question and image embedding.

\begin{figure*}[t]
  \centering
  \includegraphics[width=\linewidth]{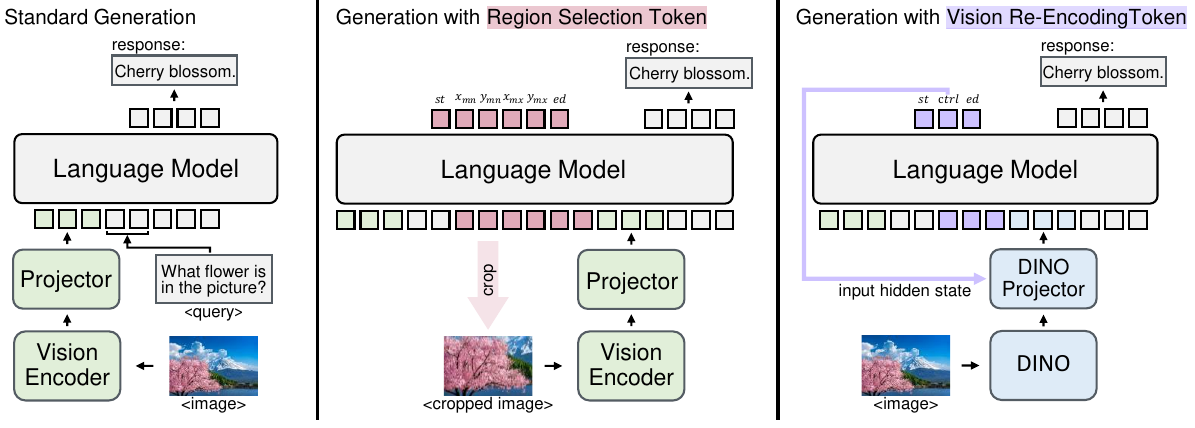}
   \caption{In a standard MLLM generation process, the model directly outputs an response based on the input image and query. However, an MLLM equipped with Visual Perception Tokens can first generate special tokens that trigger additional perception processes before responding. If the MLLM outputs a Region Selection Token, the original image is cropped and reprocessed through the visual encoder. The MLLM then bases its answer on two sets of visual embeddings: the first set contains the global embeddings from the original image, and the second set contains the local embeddings from the cropped image. If the MLLM outputs a DINO Feature Token, the DINO features of the image are used to supplement the original CLIP-based features. Additionally, besides the DINO features, the hidden state of the DINO Feature Token is also input to the projector as a condition to control which features are ultimately passed to the language model.}
   \label{fig:forward}
\end{figure*}

\begin{table*}[t]
    \centering
    \resizebox{\linewidth}{!}{%
\begin{tabular}{@{}lll@{}}
\toprule
\multicolumn{1}{c}{\textbf{Function}}                   & \multicolumn{1}{c}{\textbf{Task}} & \multicolumn{1}{c}{\textbf{Dataset}}                                           \\ \midrule
\multirow{2}{*}{Training Region Selection Token \textcolor{darkred}{(229k)}} & Text/OCR-Related VQA \textcolor{darkred}{(84k)}        & DocVQA \textcolor{darkred}{(33k)}, TextVQA \textcolor{darkred}{(19k)}, TextCaps \textcolor{darkred}{(32k)}                                    \\ \cmidrule(l){2-3} 
                                                        & Spatial Reasoning \textcolor{darkred}{(145k)}          & VSR \textcolor{darkred}{(3k)}, GQA \textcolor{darkred}{(10k)}, OpenImage \textcolor{darkred}{(43k)}                                           \\ \midrule
\multirow{2}{*}{Training DINO Feature Token \textcolor{darkpurple}{(293k)}}     & General VQA \textcolor{darkpurple}{(283k)}                & LLaVA  Instruction Tuning COCO   \textcolor{darkpurple}{(212k)}, LLaVA  Instruction Tuning GQA   \textcolor{darkpurple}{(71k)} \\ \cmidrule(l){2-3} 
                                                        & Fine-Grained VQA \textcolor{darkpurple}{(10k)}            & CUB-200-2011 \textcolor{darkpurple}{(10k)}                                                            \\ \midrule
Preserving Instruction Following Ability   \textcolor{darkgray}{(307k)}       & General VQA \textcolor{darkgray}{(307k)}                & Remaining Samples in LLaVA  Instruction   Tuning \textcolor{darkgray}{(307k)}                            \\ \bottomrule
\end{tabular}%
    }
    \caption{Composition of the training dataset. Our complete training dataset includes the data for training Visual Perception Tokens discussed in \cref{sec:data} and the remaining portion of the LLaVA-1.5 finetuning dataset. In total, the training dataset comprises 829k samples.}
    \label{tab:data}
\end{table*}
                                               
\section{MLLM with Visual Perception Token}
\subsection{Architecture}
When the Vision Re-Encoding Token triggers re-encoding, we use either an additional DINO or SAM model or the MLLM’s original vision encoder. In all cases, an extra projector is added to align vision features with LLM embeddings. This projector is a cross-attention module that takes the hidden states of the $<$Re-Encoding\_Control$>$ token as the keys and values, and the vision features as the query. This design enables the hidden states of the $<$Re-Encoding\_Control$>$ token to control the vision features finally input to the LLM. See \cref{fig:forward} for an illustration of the generation process with different types of Visual Perception Tokens. Below, we specifically describe the forward process in the modified MLLM.

An MLLM typically includes a LLM, a vision encoder  $f_v$, and a projector  $g_v$  that connects $f_v$ and the LLM. For an input image  $\bm{x}$ , the image features encoded by  $f_v$  are denoted as  $\bm{z} = f_v(\bm{x})$. After alignment through  $g_v$ , the resulting image embeddings can be represented as  $\bm{h} = g_v(\bm{z})$. These image embeddings are then concatenated with text embeddings to form the input for the LLM.
When the LLM outputs Region Selection Tokens, the cropped image  $\bm{x}'$  is reprocessed through the original vision encoder and projector, resulting in  $\bm{h}' = g_v(f_v(\bm{x}'))$, which is appended to the previous image and text embeddings as input to the LLM.

\begin{table*}[t]
    \centering
    \resizebox{\linewidth}{!}{%
\begin{tabular}{lcccccc|c}
\toprule
\multicolumn{1}{c}{}                        &                                  & \multicolumn{3}{c|}{Visual Reasoning}                                                         & \multicolumn{2}{c|}{General VQA}                                             & Fine-Grained VQA          \\ \cmidrule(l){3-8} 
\multicolumn{1}{c}{\multirow{-2}{*}{Model}} & \multirow{-2}{*}{Max Resolution} & GQA            & OpenImage      & \multicolumn{1}{c|}{VSR}                                    & LLaVA Instruction Tuning                                    & Flickr*        & CUB Birds                 \\ \midrule
Qwen2-VL-2B                                 & 224                              & 0.448          & 0.413          & \multicolumn{1}{c|}{0.561}                                  & 0.655                                                       & 0.521          & 0.621                     \\
Qwen2-VL-7B                                 & 224                              & 0.464          & 0.442          & \multicolumn{1}{c|}{0.632}                                  & 0.703                                                       & 0.524          & 0.697                     \\
\rowcolor{fullgreen!20} 
Qwen2-VL-2B-VPT (DINO)                      & 224                              & \textbf{0.606} & \textbf{0.842} & \multicolumn{1}{c|}{\cellcolor{fullgreen!20}\textbf{0.657}} & \textbf{0.705}                                              & \textbf{0.558} & \textbf{0.892}            \\ \midrule
Qwen2-VL-2B                                 & 512                              & 0.487          & 0.418          & \multicolumn{1}{c|}{0.580}                                  & 0.728                                                       & 0.557          & 0.652                     \\
Qwen2-VL-7B                                 & 512                              & 0.569          & 0.456          & \multicolumn{1}{c|}{0.641}                                  & 0.780                                                       & 0.636 & 0.687                     \\
\rowcolor{fullgreen!20} 
Qwen2-VL-2B-VPT (DINO)                      & 512                              & 0.625 & 0.872 & \multicolumn{1}{c|}{\cellcolor{fullgreen!20}0.738} & 0.797                                              & 0.663          & 0.898            \\ 
\rowcolor{fullgreen!20} 
Qwen2-VL-2B-VPT (DINO, Free Choice)         & 512                              & \textbf{0.635} & 0.874 & \multicolumn{1}{c|}{\cellcolor{fullgreen!20}0.733} & 0.802                                              & 0.705          & 0.911            \\ 
\rowcolor{fullgreen!20} 
Qwen2-VL-2B-VPT (CLIP)                      & 512                              & 0.621 & 0.872 & \multicolumn{1}{c|}{\cellcolor{fullgreen!20}0.746} & 0.791                                              & 0.660          & 0.913            \\ 
\rowcolor{fullgreen!20} 
Qwen2-VL-2B-VPT (SAM)                       & 512                              & 0.617 & 0.876 & \multicolumn{1}{c|}{\cellcolor{fullgreen!20}0.746} & 0.796                                              & 0.657          & 0.905            \\ 
\rowcolor{fullred!20} 
Qwen2-VL-7B-VPT (CLIP)                      & 512                              & 0.633 & \textbf{0.878} & \multicolumn{1}{c|}{\cellcolor{fullred!20}\textbf{0.790}} & \textbf{0.831}                                              & \textbf{0.680}          & \textbf{0.921}            \\ \bottomrule\toprule
                                            &                                  & \multicolumn{4}{c|}{Text/OCR Related VQA}                                                                                                                   & Hallucination  &                           \\ \cmidrule(lr){3-7}
\multicolumn{1}{c}{\multirow{-2}{*}{Model}}                    & \multirow{-2}{*}{Max Resolution} & DocVQA         & TextVQA        & TextCaps                                                    & \multicolumn{1}{c|}{DUDE*}                                  & POPE*          & \multirow{-2}{*}{Average} \\ \midrule
Qwen2-VL-2B                                 & 224                              & 0.051          & 0.383          & 0.390                                                       & \multicolumn{1}{c|}{0.063}                                  & 0.821          & 0.448                     \\
Qwen2-VL-7B                                 & 224                              & 0.063          & 0.421          & 0.431                                                       & \multicolumn{1}{c|}{0.095}                                  & 0.827          & 0.482                     \\
\rowcolor{fullgreen!20} 
Qwen2-VL-2B-VPT (DINO)                      & 224                              & \textbf{0.125} & \textbf{0.537} & \textbf{0.466}                                              & \multicolumn{1}{c|}{\cellcolor{fullgreen!20}\textbf{0.103}} & \textbf{0.843} & \textbf{0.576}            \\ \midrule
Qwen2-VL-2B                                 & 512                              & 0.301          & 0.765          & 0.710                                                       & \multicolumn{1}{c|}{0.253}                                  & 0.847          & 0.572                     \\
Qwen2-VL-7B                                 & 512                              & 0.360          & 0.816 & 0.732                                              & \multicolumn{1}{c|}{0.322}                                  & 0.866          & 0.624                     \\
\rowcolor{fullgreen!20} 
Qwen2-VL-2B-VPT (DINO)                      & 512                              & 0.573 & 0.860          & 0.766                                                       & \multicolumn{1}{c|}{\cellcolor{fullgreen!20}0.430} & 0.893 & 0.738            \\ 
\rowcolor{fullgreen!20} 
Qwen2-VL-2B-VPT (DINO, Free Choice)         & 512                              & 0.576 & 0.861          & 0.758                                                       & \multicolumn{1}{c|}{\cellcolor{fullgreen!20}0.438} & 0.950 & 0.749            \\ 
\rowcolor{fullgreen!20} 
Qwen2-VL-2B-VPT (CLIP)                      & 512                              & 0.567 & 0.856          & 0.770                                                       & \multicolumn{1}{c|}{\cellcolor{fullgreen!20}0.433} & 0.887 & 0.738            \\ 
\rowcolor{fullgreen!20} 
Qwen2-VL-2B-VPT (SAM)                       & 512                              & 0.558 & 0.858          & 0.750                                                       & \multicolumn{1}{c|}{\cellcolor{fullgreen!20}0.431} & 0.894 & 0.735            \\ 
\rowcolor{fullred!20} 
Qwen2-VL-7B-VPT (CLIP)                      & 512                              & \textbf{0.658} & \textbf{0.906}          & \textbf{0.788}                                                       & \multicolumn{1}{c|}{\cellcolor{fullred!20}\textbf{0.532}} & \textbf{0.903} & \textbf{0.773}            \\ \bottomrule
\end{tabular}%
    }
    \caption{Performance comparison of MLLMs with and without Visual Perception Tokens. Datasets marked with ``*'' are not used in the training process. The best performance is highlighted in \textbf{bold}. A 2B model with Visual Perception Tokens can even outperform the 7B model without Visual Perception Tokens.}
    \label{tab:main}
\end{table*}

Let  $f_D$  denote the vision model for re-encoding,  $g_D$  the projector between  $f_D$ and the LLM, and  $\bm{h}_{DC} \in \mathbb{R}^{1 \times d_h}$  denote the hidden state of the $<$Re-Encoding\_Control$>$ token, where  $d_h$  is the hidden size of the LLM. When the LLM outputs $<$Re-Encoding\_Control$>$ token, the original image is input  into the $f_D$, yielding features  $\bm{z}_D = f_D(\bm{x}) \in \mathbb{R}^{N \times d_z}$ , where  $N$  is the sequence length of the re-encoded image features and  $d_z$  is the hidden size of $f_D$. The features  $\bm{z}_D$  and  $\bm{h}_{DC}$  are then input to the projector, resulting in the embeddings  $\bm{h}_D = g_D(\bm{z}_D, \bm{h}_{DC}) \in \mathbb{R}^{N \times d_h}$. $\bm{h}_D$ is then concatenated with the previous image and text embeddings as input to the LLM.

\subsection{Training Data for Visual Perception Token}\label{sec:data}
We constructed the training dataset for Visual Perception Token based on the datasets from \cite{llava15} and \cite{shao2024visual}. Our training data covers four types of tasks: Text/OCR-Related VQA, Spatial Reasoning, General VQA, and Fine-Grained VQA. The Text/OCR-Related VQA and Spatial Reasoning tasks are used to create training samples for Region Selection Token. The General VQA and Fine-Grained VQA tasks are used to construct training samples for Vision Re-Encoding Tokens. Please see \cref{tab:data} for the composition of training dataset.

\textbf{Samples with Region Selection Tokens}. We used the DocVQA~\cite{docvqa},  TextVQA~\cite{textvqa}, and TextCaps~\cite{textcaps} datasets to build training data for Text/OCR-Related tasks. DocVQA images include photos of documents such as books and invoices, while TextVQA and TextCaps images feature natural scenes such as billboards and store signs. Both datasets involve questions requiring reasoning about text within images. For these datasets, we use the text regions' bounding boxes obtained in \cite{shao2024visual}, and convert the bounding boxes into Region Selection Token sequences as described in \cref{sec:31}. The training data for Spatial Reasoning tasks included images from the VSR~\cite{vsr}, GQA~\cite{hudson2019gqa}, and OpenImage~\cite{openimage} datasets, which contain real-world scenes and questions about the spatial relationships between objects. We used the filtered bounding boxes and question-answer pairs from \cite{shao2024visual} and converted the bounding boxes into the corresponding Region Selection Tokens.

\textbf{Samples with Vision Re-Encoding Tokens}. For General VQA tasks, we used data from COCO~\cite{coco} and GQA~\cite{hudson2019gqa} as provided by LLaVA 1.5~\cite{llava15}. We inserted an additional MLLM response with Vision Re-Encoding Tokens between the original question and response. For Fine-Grained VQA tasks, we used the CUB-200-2011~\cite{WahCUB_200_2011} dataset, which includes images of 200 bird species and annotations about detailed attributes of these birds. We used the question-answer pairs generated in \cite{shao2024visual} and inserted an MLLM response with Vision Re-Encoding Tokens between the questions and answers.

\section{Experiments}
\subsection{Main Results}

We used Qwen2-VL-2B or Qwen2-VL-7B~\cite{Qwen2VL} as the base MLLM model and DINOv2~\cite{oquab2024dinov} or SAM~\cite{sam} as the additional vision feature extractor. We use CLIP to denote the case when the original vision branch is used in the re-encoding process. The 2B models are fully fine-tuned, while the 7B model are tuned using LoRA.
The datasets described in \cref{sec:data} were used to finetune the model. 
We then compared its performance with the original Qwen2-VL-2B and Qwen2-VL-7B models.

We evaluated the performance of Qwen2-VL-2B-VPT on the test sets of the datasets used in \cref{sec:data}. When an official test split was available, we used it directly. When no official test split was provided, we randomly split the data into training and testing sets and filter out any images that appeared in both sets to prevent data leakage. Additionally, we also conducted evaluations on Flickr~\cite{flickr}, DUDE~\cite{dude}, and POPE~\cite{pope} datasets, which were not used in training, to assess the generalization capability of our method. 

Following established practices~\cite{mmvet,llava15}, we used GPT-4o (2024-08-06) to evaluate the alignment between the model's responses and the ground truth for each question. A higher score indicates a higher degree of matching, with 0 representing no match and 1 indicating a perfect match. We reported the average matching scores for each dataset. The prompts used in the evaluation are included in the supplementary material.
\begin{figure*}[t]
  \centering
  \includegraphics[width=\linewidth]{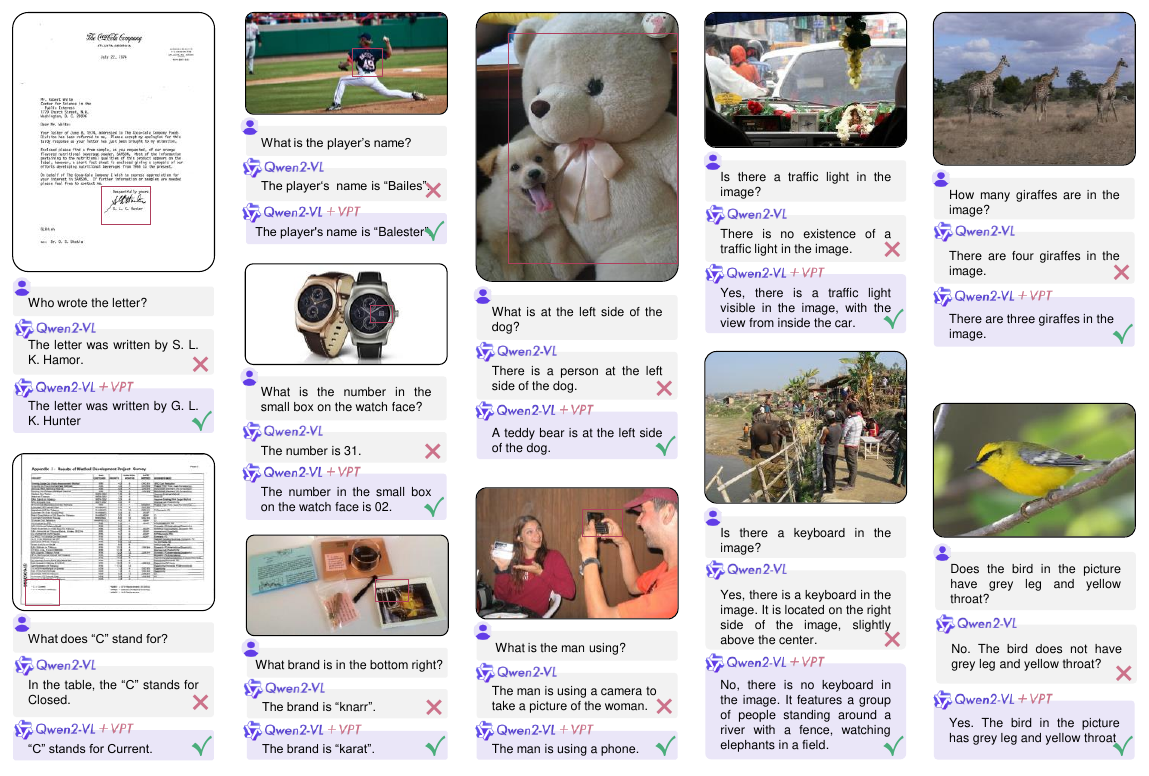}
   \caption{Examples collected from the testing sets. The responses were generated by the 7B model and the 2B+VPT model. During the generation process, if Region Selection Tokens were utilized, the region selected by these tokens are highlighted with red boxes in the images. For additional examples, please refer to the supplementary material. 
   }
   \label{fig:examples}
   \vspace{-1em}
\end{figure*}

The experimental results are presented in \cref{tab:main}. In terms of overall average performance, 2B model with Visual Perception Tokens outperforms the 7B model without Visual Perception Tokens by a significant margin. 
A more detailed analysis is as follows:
(1) The trend of using different vision encoders for re-encoding remains consistent, with the 2B model enhanced by Visual Perception Tokens outperforming the 7B model.  
(2) We implemented two methods for triggering Visual Perception Tokens. The first method explicitly enforces the use of either Region Selection or DINO Feature Tokens across the entire dataset, aligning with real-world scenarios where users may require explicit control. This corresponds to most results in \cref{tab:main}.  
The second allows the model to autonomously decide whether to use Perception Tokens and which type to apply, corresponding to the ``Free Choice'' results in \cref{tab:main}. This ``Free Choice'' mechanism achieves even better performance than the first approach.
(3) The Visual Perception Token shows a clear advantage in Visual Reasoning and Fine-Grained VQA tasks, with improvements of 0.161 and 0.207 over the 7B model, respectively. However, in some datasets for General VQA and Text/OCR-related VQA tasks, the 2B model with Visual Perception Tokens only performs comparably to the 7B model, without a significant performance gain.
(4) The Visual Perception Token remains effective in zero-shot settings. During evaluation, we included three datasets that were not used in training. On these datasets, the 2B-VPT model still outperforms or matches the performance of the 7B model.
(5) The Visual Perception Token is effective for both high- and low-resolution images. On average, when using low-resolution images as input, the 2B+VPT model improves by 0.094 over the 7B model, while for high-resolution images, it shows an improvement of 0.084 over the 7B model.

In \cref{fig:examples}, we show examples demonstrating the effectiveness of Visual Perception Tokens.
Visual Perception Tokens are particularly effective for several types of queries:
The first category involves locating small regions within large documents or images. The small regions can be signatures, footnotes, page numbers, product brands, logos, or small text on clothing, which are often too small for direct OCR by MLLMs. However, Region Selection Tokens can pinpoint these areas.
The second category addresses hallucination issues. For example, when asked what is next to a dog, text-based hallucination might generate an answer like ``a person''. Region Selection Tokens mitigate such errors by grounding responses in image.
The third category involves identifying or counting objects in complex scenes. Vision Re-Encoding Tokens excel in these tasks because encoding the image twice enhances segmentation and detection.

\subsection{Discussion on Region Selection Token}
In this subsection, we assess the necessity of introducing Region Selection Tokens and examine whether directly using bounding boxes is an effective approach for indicating selected regions. Additionally, we conduct an ablation study on the granularity parameter $k$.

Due to the cost of experiments, we did not train models on all the datasets used in the main experiment for ablation. Instead, we focused on training using the DocVQA, TextVQA, and TextCaps datasets and evaluated on their test splits. This choice was made because, for text/OCR-related VQA tasks, the query-related region is typically small, and the images often contain complex objects that distract the MLLM.  Therefore, the quality of region selection substantially affects the final VQA performance. 
We trained five models in total. The first four models employed Region Selection Tokens with $k$ values of 4, 8, 16, and 32, where a larger $k$ indicates finer granularity in the image partition. The fifth model directly used bounding boxes, as described in \cite{shao2024visual}. The results are presented in \cref{tab:ablation_k}. 

The findings are twofold: (1) Region Selection Tokens prove to be more effective than bounding boxes. As shown in the results, models with $k=8$ and $k=16$ significantly outperformed model that used bounding boxes as region indicators. We also observed that when MLLMs utilized bounding boxes, they generated many invalid bounding boxes, such as those where the height or width exceeded the original image dimensions.
(2) There is an optimal granularity $k$. In our experiments, $k=8$ generally yielded the best performance. When $k$ is too small, the cropped region remains large and can still contain complex scenes or multiple objects, which fails to guide the model's attention effectively. Conversely, when $k$ is too large, more new tokens are required, increasing the learning difficulty for the MLLM. Moreover, overly fine-grained region descriptions do not contribute to VQA performance. These two factors together leads to an overall decline in performance.

\subsection{Discussion on Vision Re-Encoding Token}
In this subsection, we validate the effectiveness of the Vision Re-Encoding Token and conduct an ablation on it. 

To verify the effectiveness of the Vision Re-Encoding Token, \cref{tab:validation_dino_token} presents the performance comparison between two model variants. The model with control information (last row in the table) includes the hidden state of the $<$Re-Encoding\_Control$>$ token as an input to the projector, enabling fine-grained control over the embeddings fed into the LLM. 
In contrast, the model without control information  (middle row in the table) uses directly the re-encoded image feature as input to the LLM, after aligning its dimension with the embedding dimension of the LLM. In this experiment, DINO-v2 is used as the additional vision encoder.
The experimental results show that the additional control information carried by the $<$Re-Encoding\_Control$>$ token significantly enhances model performance.

\begin{table}[t]
    \centering
    \resizebox{\linewidth}{!}{%
    \begin{tabular}{llccc}
    \toprule
      && DocVQA & TextVQA & TextCaps \\ \midrule
    \multirow{4}{*}{\makecell{Region\\Selection\\Token}} & $k=4$                 & 0.251  & 0.599   & 0.521   \\
    &$k=8$                  & \textbf{0.369}  & 0.686   & \textbf{0.578}   \\
    &$k=16$                 & 0.307  & \textbf{0.690}   & 0.569   \\
    &$k=32$                 & 0.264  & 0.634   & 0.539   \\ \midrule
    \multicolumn{2}{c}{Bounding Box}          & 0.219  & 0.620   & 0.547   \\ \bottomrule
    \end{tabular}%
    }
    \caption{Ablation on the parameter $k$, which controls the granularity of the Region Selection Token and whether Bounding Boxes should be used directly.}
    \label{tab:ablation_k}
\end{table}

\begin{table}[t]
    \centering
    \resizebox{\linewidth}{!}{%
    \begin{tabular}{c@{$\quad$}c@{$\quad$}c@{$\quad$}c}
    \toprule
    Control Info & CUB Bird & LLaVA COCO                      & LLaVA GQA                       \\ \midrule
    \xmark                                            & 0.7459                        & 0.6226                    & 0.4354                    \\
    \cmark                                            & \textbf{0.8943}                        & \textbf{0.7804}                    & \textbf{0.7908}                    \\ \bottomrule
    \end{tabular}%
    }
    \caption{Validation of the designed Vision Re-Encoding Token's ability to convey valuable control information.}
    \label{tab:validation_dino_token}
\end{table}

\begin{table}[t]
    \centering
    \resizebox{\linewidth}{!}{%
    \begin{tabular}{@{}l@{}c@{$\ \ $}c@{$\ \ $}c@{$\ \ $}c@{$\ \ $}c@{}}
    \toprule
    Method                          & \makecell{Mask\\Modeling} & \makecell{No. of Ctrl\\Tokens} & \makecell{CUB\\Bird} & \makecell{LLaVA\\COCO} & \makecell{LLaVA\\GQA} \\ \midrule
    Qwen2-VL-2B                     &                        & 0                       & 0.882    & 0.718      & 0.734     \\
    2B+VPT (DINO)                   &                        & 1                       & 0.892    & 0.730      & 0.758     \\
    2B+VPT (DINO)                   &                        & 2                       & 0.896    & 0.731      & 0.761     \\
    2B+VPT (DINO)                   &                        & 4                       & 0.876    & 0.715      & 0.759     \\
    2B+VPT (DINO)                   & \cmark                      & 1                       & 0.901    & 0.739      & 0.763     \\
    $\quad$+Tune Projector & \cmark                      & 1                       & \textbf{0.918}    & \textbf{0.755}      & 0.769     \\
    2B+VPT (CLIP)                   & \cmark                      & 1                       & 0.902    & 0.728      & 0.770     \\
    $\quad$+Tune Projector & \cmark                      & 1                       & 0.909    & 0.734      & \textbf{0.775}     \\ \bottomrule
    \end{tabular}%
    }
    \caption{Ablation on the number of Vision Re-Encoding Token. On one hand, with Mask Modeling the performance is improved. On the other hand, the results do not suggest a significant improvement when using more $<$\text{Re-Encoding\_Control}$>$ tokens.}
    \label{tab:ablation_num_dino_ctrl}
    \vspace{-1em}
\end{table}

Next, we conduct an ablation study on the number of $<$Re-Encoding\_Control$>$ tokens and mask modeling. Considering the computational cost, in these experiments, we trained  models using only the training data from the CUB Bird dataset and LLaVA Instruction Tuning data from COCO and GQA. 
Results are presented in \cref{tab:ablation_num_dino_ctrl} and indicate the following conclusions.
First, increasing the number of $<$Re-Encoding\_Control$>$ tokens from 1 to 2 provides limited performance gains, while increasing to 4 leads to a decline. This is due to the projector`s lightweight design, making it susceptible to over-parameterization and overfitting.  
Second, introducing masked modeling improves the performance. Further experiments show that fine-tuning the linear layer connecting the hidden states of $<$Re-Encoding\_Control$>$ in the projector for 1 extra epoch further improves performance, confirming that these hidden states encode valuable control information.
\section{Conclusion}

In this work, we propose Visual Perception Tokens to enable MLLM to autonomously controls its visual perception process. MLLM can generate these Visual Perception Tokens in a manner similar to generating natural language and use them to convey information to the visual perception process. 
Experiments validate the effectiveness of Visual Perception Token over various tasks.

\clearpage

\onecolumn
\renewcommand{\thefigure}{S\arabic{figure}}
\renewcommand{\thetable}{S\arabic{table}}
\renewcommand{\thesection}{S\arabic{section}}
\setcounter{figure}{0}
\setcounter{table}{0}
\setcounter{section}{0}
{
    \centering
    \Large
    \textbf{Introducing Visual Perception Token into Multimodal Large Language Model\\\textit{- Supplementary Material -}}\\
    \vspace{1.0em}
}

\section{Implement Details}
\subsection{Training Details}
Our training process consists of two phases: alignment and finetuning. 
The alignment stage aligns the additional vision features with the LLM embeddings. If the original vision encoder is used for re-encoding, the alignment stage is omitted. We use the same image-text pair data for the LLaVA 1.5 alignment, and only use the additional vision branch as the LLM's input. During training, all components except the projector are frozen. In this phase, we train the model for 1 epoch with a learning rate of 2e-3 and a batch size of 128.
The second finetuning stage allows the model to learn to output the correct Region Selection Tokens and to transmit information through the Vision Re-Encoding Tokens. We finetune the model using our constructed  dataset, as well as remaining samples from the LLaVA 1.5 finetuning dataset that were not included in our dataset. In this stage, all components except the original visual encoder and the additional vision encoder are unfrozen. In this phase, we train the model for 1 epoch with a learning rate of 2e-5 and a batch size of 256. For both the first and the second phase, we use AdamW optimizer. The experiments are deployed on 8 A100 GPU. The total training time is about 20 hours. For the 7B model, the rank of the LoRA is set to 512.

\subsection{Evaluation Prompt}
Following established practices~\cite{mmvet,llava15}, we used GPT-4o (2024-08-06) to evaluate the alignment between the model's responses and the ground truth for each question. We use the evaluation prompt in \cite{shao2024visual}.
\begin{tcolorbox}[parbox=false,colback=fullgreen!10, colframe=fullgreen!50, title=Evaluation Prompt, coltitle=black]
You are responsible for proofreading the answers, you need to give a score to the model's answer by referring to the standard answer, based on the given question. The full score is 1 point and the minimum score is 0 points. Please output the score in the form 'score: $<$score$>$'. The evaluation criteria require that the closer the model's answer is to the standard answer, the higher the score.

\noindent Question:  $<$question$>$

\noindent Ground Truth:  $<$ground truth$>$

\noindent Answer:  $<$answer$>$
\end{tcolorbox}

\subsection{Template of the Training Data Examples}
Here, we show the format of our training examples. 
The training example for the Region Selection Token is essentially the same as the samples used in \cite{shao2024visual}, except that the method for representing regions has changed from bounding boxes to region tokens.  
The training example for the Vision Re-Encoding Token is almost identical to the data in the original LLava~\cite{llava} fine-tuning dataset, with the only difference being the insertion of an additional round of dialogue between the original question and answer. This added dialogue includes the Vision Re-Encoding Token.  
\begin{tcolorbox}[parbox=false,colback=fullgreen!10, colframe=fullgreen!50, title=Template of Training Example for Region Selection Token, coltitle=black]
\textbf{User}: $<$image$>$ $<$question$>$ Please identify the region that can help you answer the question better, and then answer the question.

\noindent\textbf{Assistant}: $<$Region\_Selection\_Start$>$ $<$x\_min$>$ $<$y\_min$>$ $<$x\_max$>$ $<$y\_max$>$ $<$Region\_Selection\_End$>$.

\noindent\textbf{User}: $<$image$>$

\noindent\textbf{Assistant}: $<$ground truth$>$
\end{tcolorbox}
\begin{tcolorbox}[parbox=false,colback=fullgreen!10, colframe=fullgreen!50, title=Template of Training Example for Vision Re-Encoding Token, coltitle=black]
\textbf{User}: $<$image$>$ $<$question$>$ Please require additional perception features, and then answer the question.

\noindent\textbf{Assistant}: $<$Re-Encoding\_Start$>$  $<$Re-Encoding\_Control$>$ $<$Re-Encoding\_End$>$.

\noindent\textbf{User}: $<$image$>$

\noindent\textbf{Assistant}: $<$ground truth$>$
\end{tcolorbox}

The training for the free-choice experiment differs from other experiments only in the sample template. For the free-choice experiment, we removed the additional prompt from the questions. The training sample template is as follows.

\begin{tcolorbox}[parbox=false,colback=fullgreen!10, colframe=fullgreen!50, title=Template of Training Example for Region Selection Token (Free Choice), coltitle=black]
\textbf{User}: $<$image$>$ $<$question$>$

\noindent\textbf{Assistant}: $<$Region\_Selection\_Start$>$ $<$x\_min$>$ $<$y\_min$>$ $<$x\_max$>$ $<$y\_max$>$ $<$Region\_Selection\_End$>$.

\noindent\textbf{User}: $<$image$>$

\noindent\textbf{Assistant}: $<$ground truth$>$
\end{tcolorbox}
\begin{tcolorbox}[parbox=false,colback=fullgreen!10, colframe=fullgreen!50, title=Template of Training Example for Vision Re-Encoding Token (Free Choice), coltitle=black]
\textbf{User}: $<$image$>$ $<$question$>$

\noindent\textbf{Assistant}: $<$Re-Encoding\_Start$>$  $<$Re-Encoding\_Control$>$ $<$Re-Encoding\_End$>$.

\noindent\textbf{User}: $<$image$>$

\noindent\textbf{Assistant}: $<$ground truth$>$
\end{tcolorbox}

\section{Supplementary Experiments}
We conducted experiments on the MME~\cite{mme} and MM-Bench~\cite{mmb} benchmarks without using the Visual Perception Token, allowing the model to generate answers directly. This assessed the impact of our fine-tuning on general benchmarks. Results in \cref{tab:benchmark} show that our model does not cause degeneration and even improves performance on these benchmarks. 
\begin{table*}[t]
    \centering
    \begin{tabular}{@{}lcccc@{}}
    \toprule
    \multirow{2}{*}{} & \multicolumn{2}{c}{MME}      & \multicolumn{2}{c}{MMB}         \\ \cmidrule(l){2-5} 
                      & Cognition     & Perception   & en             & cn             \\ \midrule
    Qwen2-VL-2B       & 1434          & 280          & 78.20          & 77.30          \\
    Qwen2-VL-7B       & 1664          & 335          & 78.70          & \textbf{83.30} \\
    2B+VPT (DINO)     & 1511          & 274          & 79.11          & 76.64          \\
    2B+VPT (CLIP)     & 1510          & 273          & 79.53          & 77.41          \\
    2B+VPT (SAM)      & 1475          & 270          & 80.22          & 76.99          \\
    7B+VPT (CLIP)     & \textbf{1706} & \textbf{336} & \textbf{83.80} & \textbf{83.30} \\ \bottomrule
    \end{tabular}
    \caption{Performance comparison of MLLMs with and without Visual Perception Tokens on MME and MMBench Benchmarks.}
    \label{tab:benchmark}
\end{table*}

To verify the advantage of the Region Selection Token over direct BBox prediction, we compared the predicted regions with ground truth using IoU and Intersection over Ground Truth (IoGT), defined as:  
$$(\text{IoGT} = \frac{\text{Area of } (GT \cap \text{Pred})}{\text{Area of } GT}).$$ Results in \cref{tab:iou} show that Region Selection Token significantly outperforms direct BBox prediction in accuracy.  
\begin{table*}[t]
    \centering
    \begin{tabular}{@{}llccc@{}}
    \toprule
                                                    & \multicolumn{1}{c}{Metric} & DocVQA & TextVQA & TextCap \\ \midrule
    \multirow{2}{*}{Directly   Predicting BBox}     & IoU                        & 0.15   & 0.26    & 0.25    \\
                                                    & IoGT                       & 0.20   & 0.28    & 0.27    \\
    \multirow{2}{*}{Using   Region Selection Token} & IoU                        & 0.26   & 0.56    & 0.50    \\
                                                    & IoGT                       & 0.38   & 0.71    & 0.66    \\ \bottomrule
    \end{tabular}
    \caption{Performance comparison of MLLMs with and without Visual Perception Tokens on MME and MMBench Benchmarks.}
    \label{tab:iou}
\end{table*}

\section{Further Examples}
\begin{figure*}[h]
  \centering
    \begin{minipage}[t]{0.45\textwidth}
        \centering
        \vspace{0pt}
        \includegraphics[width=\textwidth]{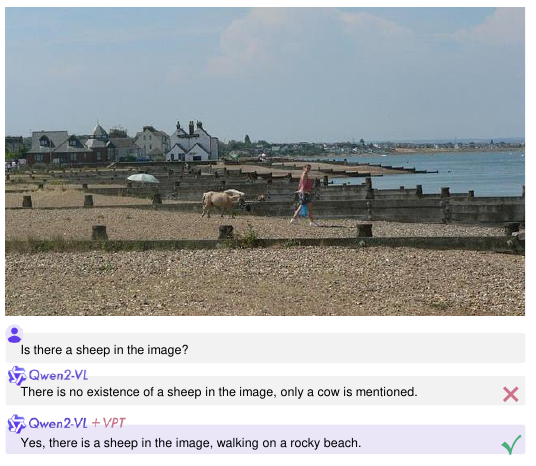}
    \end{minipage}
    \hfill
    \begin{minipage}[t]{0.45\textwidth}
        \centering
        \vspace{0pt}
        \includegraphics[width=\textwidth]{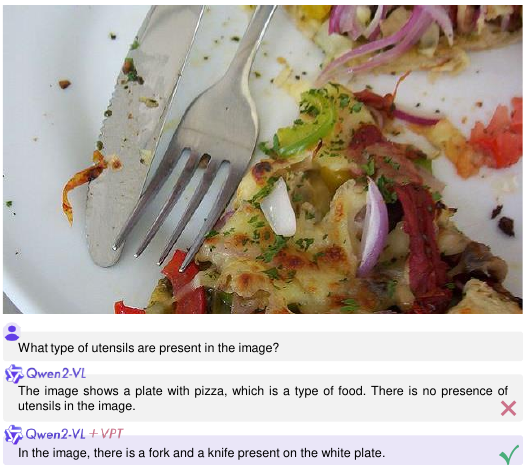}
    \end{minipage}
   \caption{This set of images demonstrates how the DINO Feature Token assists MLLMs in identifying specific objects within images. These objects are often difficult for MLLMs to recognize directly due to their small size or interference from surrounding objects.}
   \label{fig:moreexample:g3}
\end{figure*}

\begin{figure*}[h]
\vspace{0pt}
  \centering
    \begin{minipage}[t]{0.45\textwidth}
        \centering
        \vspace{0pt}
        \includegraphics[width=\textwidth]{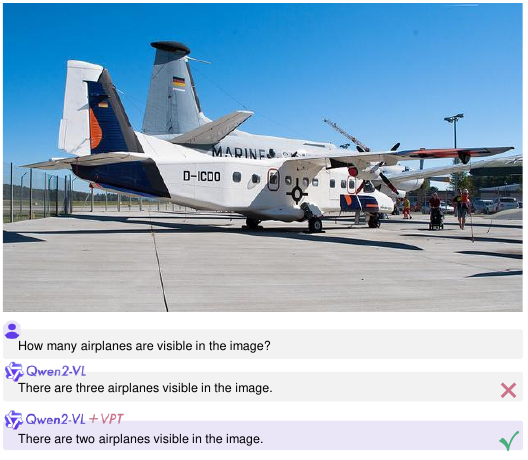}
    \end{minipage}
    \hfill
    \begin{minipage}[t]{0.45\textwidth}
        \centering
        \vspace{0pt}
        \includegraphics[width=\textwidth]{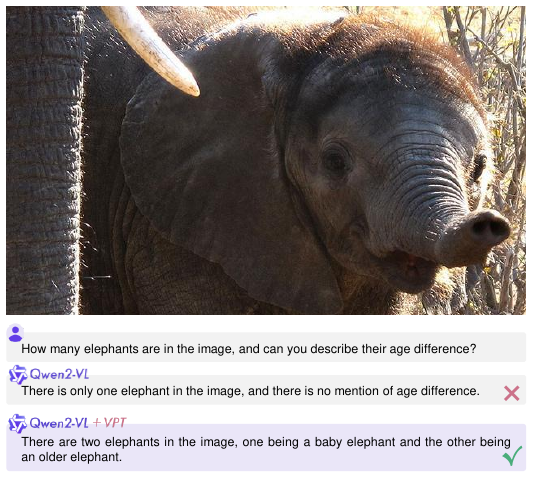}
    \end{minipage}
    \\
    \begin{minipage}[t]{0.45\textwidth}
        \centering
        \vspace{0pt}
        \includegraphics[width=\textwidth]{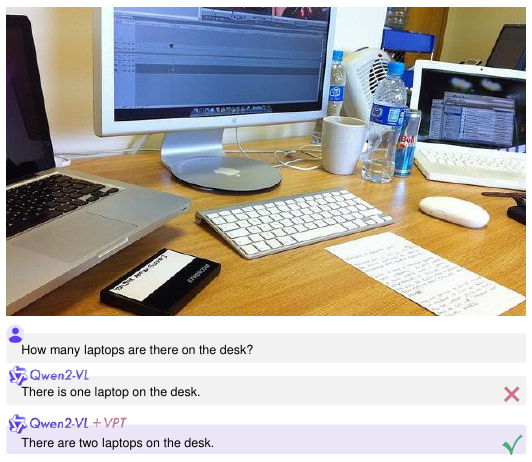}
    \end{minipage}
    \hfill
    \begin{minipage}[t]{0.45\textwidth}
        \centering
        \vspace{0pt}
        \includegraphics[width=\textwidth]{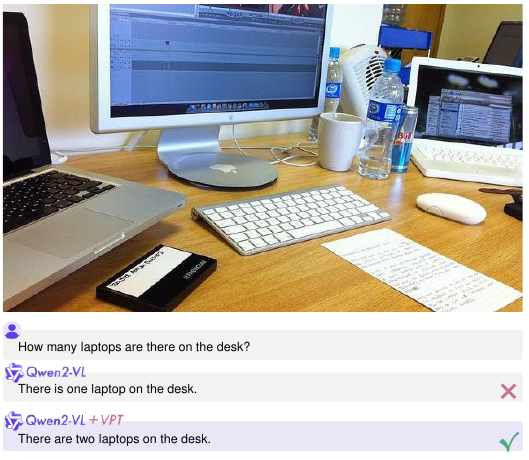}
    \end{minipage}
   \caption{This set of images illustrates how the DINO Feature Token assists MLLMs in counting the number of objects in an image. Counting has long been a significant limitation for MLLMs. By leveraging the DINO Feature, the DINO Feature Token enables precise localization of individual objects within the image, thereby improving the counting capability of MLLMs.}
   \label{fig:moreexample:g4}
\end{figure*}

\begin{figure*}[h]
  \centering
    \begin{minipage}[t]{0.45\textwidth}
        \centering
        \vspace{0pt}
        \includegraphics[width=\textwidth]{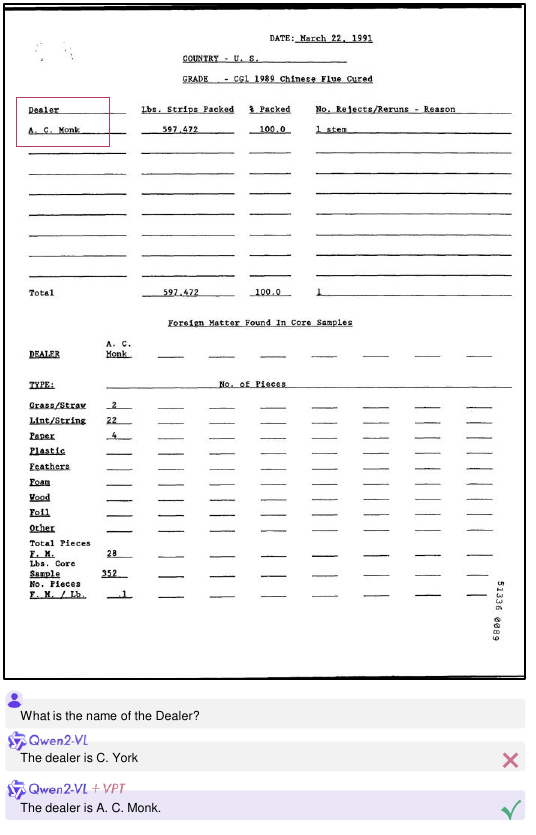}
    \end{minipage}
    \hfill
    \begin{minipage}[t]{0.45\textwidth}
        \centering
        \vspace{0pt}
        \includegraphics[width=\textwidth]{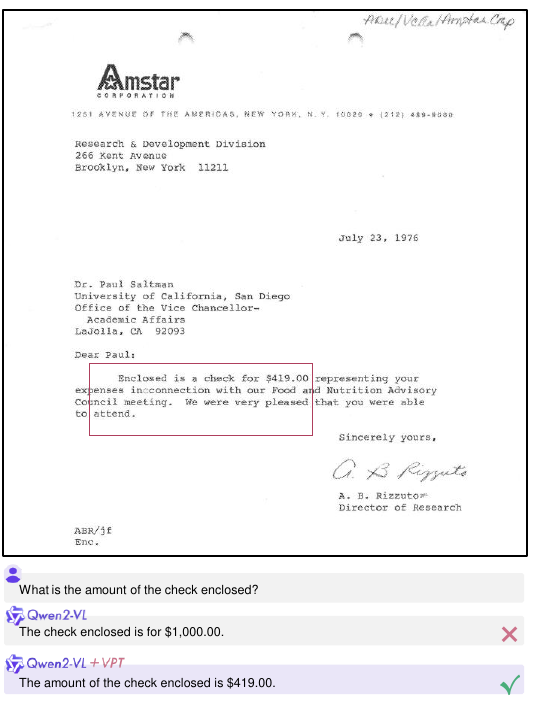}
    \end{minipage}
    \\
    \begin{minipage}[t]{0.45\textwidth}
        \centering
        \vspace{0pt}
        \includegraphics[width=\textwidth]{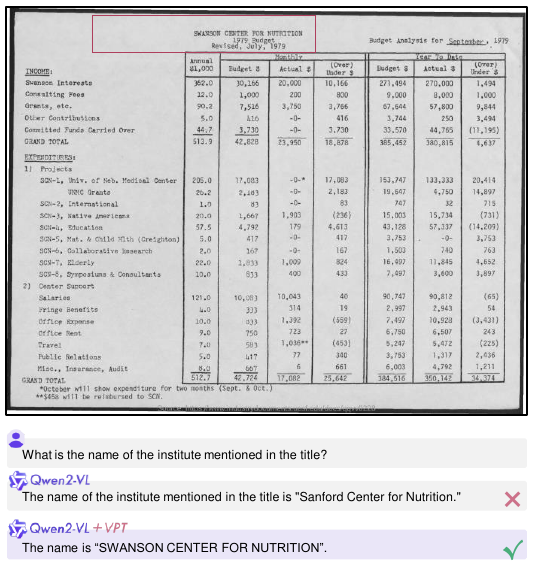}
    \end{minipage}
    \hfill
    \begin{minipage}[t]{0.45\textwidth}
        \centering
        \vspace{0pt}
        \includegraphics[width=\textwidth]{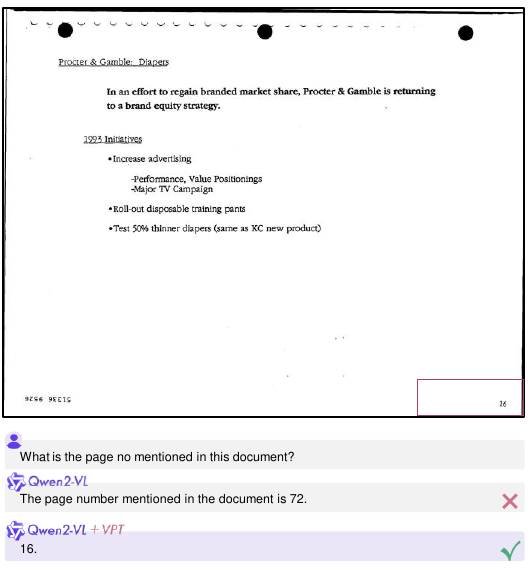}
    \end{minipage}
   \caption{This group of examples shows how the Region Selection Token aids MLLMs in understanding textual information within images by correctly identifying the corresponding regions. The image inputs primarily consist of large but structured documents, such as tables, forms, or letters.}
   \label{fig:moreexample:g1}
\end{figure*}

\begin{figure*}[h]
  \centering
    \begin{minipage}[t]{0.45\textwidth}
        \centering
        \vspace{0pt}
        \includegraphics[width=\textwidth]{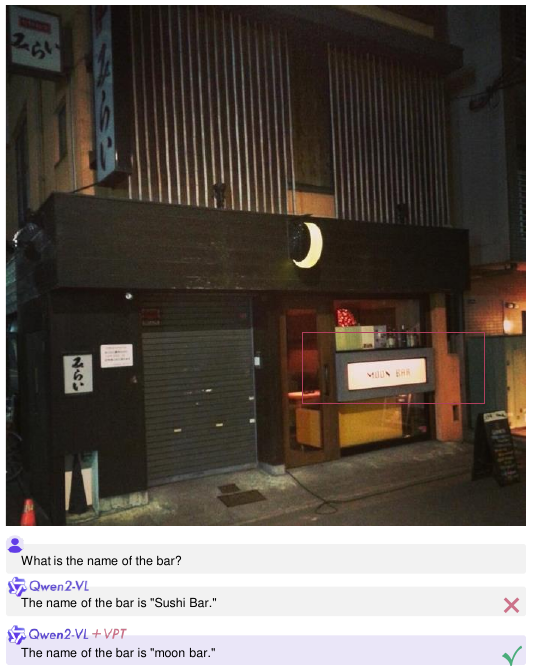}
    \end{minipage}
    \hfill
    \begin{minipage}[t]{0.45\textwidth}
        \centering
        \vspace{0pt}
        \includegraphics[width=\textwidth]{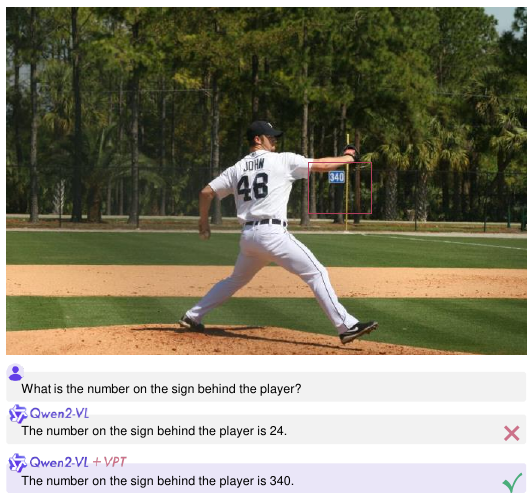}
    \end{minipage}
    \\
    \begin{minipage}[t]{0.45\textwidth}
        \centering
        \vspace{0pt}
        \includegraphics[width=\textwidth]{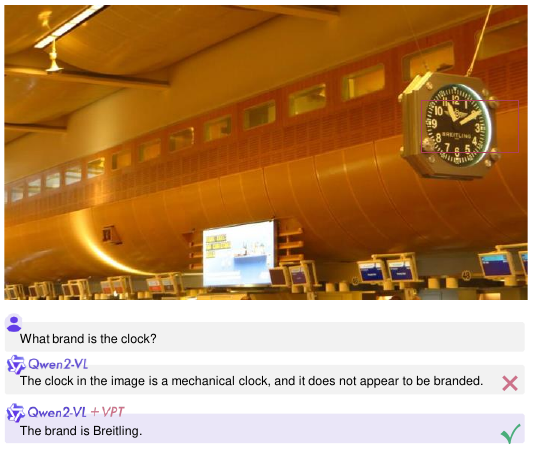}
    \end{minipage}
    \hfill
    \begin{minipage}[t]{0.45\textwidth}
        \centering
        \vspace{0pt}
        \includegraphics[width=\textwidth]{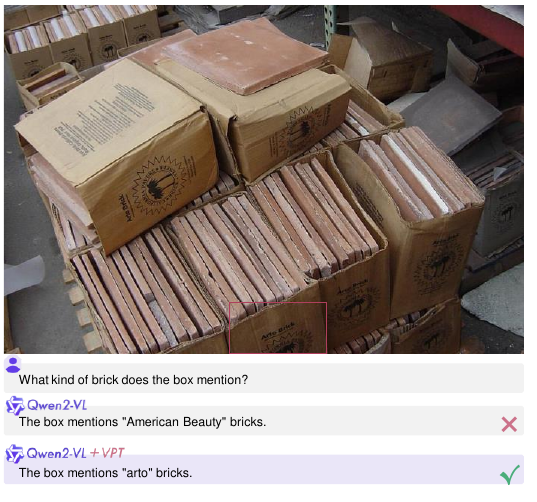}
    \end{minipage}
   \caption{This set of images illustrates how the Region Selection Token enables MLLMs to comprehend textual information within real-world scenes by accurately identifying the corresponding regions. The image inputs consist of real-world scenarios, such as signboards and trademarks, where the text occupies only a small portion of the overall scene and is highly susceptible to interference from the surrounding context.}
   \label{fig:moreexample:g2}
\end{figure*}

Here we present additional examples obtained using the visual perception token. \cref{fig:moreexample:g3,fig:moreexample:g4} include the responses generated with the Vision Re-Encoding Token. \cref{fig:moreexample:g1,fig:moreexample:g2} present the responses generated with the Region Selection Token, with the regions selected by the Region Selection Token highlighted in the images.
\textcolor{white}{empty}
\vspace{0pt}
\textcolor{white}{empty}
\textcolor{white}{empty}
\textcolor{white}{empty}

\section{Additional Related Works}
\subsection{Reasoning Token} 
In Large Language Model (LLM), there are tokens, similar to Visual Perception Token, which are designed to control the generation process of LLM. These token are termed reasoning tokens or planning token and have recently been introduced in OpenAI's o1 model~\cite{openai_reasoning_guide} and other LLMs.
For example, to enhances models' reasoning capabilities, reasoning tokens were explicitly integrated into OpenAI's o1 models to segment prompts into smaller, manageable parts, exploring multiple response strategies before generating the final output~\cite{openai_reasoning_guide}.  Similar methods aim to incorporate CoT reasoning into language models through planning tokens or distillation techniques. For example, a hierarchical generation framework using planning tokens has been proposed, embedding high-level plans at each reasoning stage with minimal parameter increase~\cite{wang2024guiding}. Moreover, techniques like Rephrase and Respond have been distilled back into models, improving efficiency and accuracy in reasoning, as demonstrated in \cite{distill2to1}.

Our work focuses on MLLMs, where we design visual perception tokens to enhance the visual perception capabilities of MLLMs, not for LLM.
Moreover, our exploration goes beyond LLM reasoning tokens. Unlike these tokens, which merely trigger specific actions and lack the ability to convey detailed instructions or rich information, we focus on designing tokens capable of transmitting nuanced control information for fine-grained visual perception.

\section{Discussion}
\textbf{Adaptability of Visual Perception Token.}
The design of the visual perception token depends on the specific visual perception method. In this paper, we use Crop and the addition of vision features as examples to introduce two types of visual perception tokens. However, our approach can be extended to other visual prompting techniques or visual encoder models, and even to LLM-agent or LLM-tool systems beyond vision.

\clearpage
{
    \small
    \bibliographystyle{ieeenat_fullname}
    \bibliography{main}
}


\end{document}